\documentclass[times,twocolumn,final,authoryear]{article}

\usepackage{graphicx}

\usepackage{amssymb}
\usepackage{latexsym}
\usepackage{amsmath}

\usepackage{url}
\usepackage{xcolor}

\newcommand{\dt}{\Delta t}
\newcommand{\s}{\mathbf{s}}
\renewcommand{\u}{\mathbf{u}}
\renewcommand{\c}{\mathbf{c}}
\newcommand{\q}{\mathbf{q}}
\renewcommand{\v}{\mathbf{v}}
\newcommand{\w}{\mathbf{\omega}}
\newcommand{\x}{\mathbf{x}}
\newcommand{\z}{\mathbf{z}}
\renewcommand{\P}{\mathtt{P}}
\newcommand{\F}{\mathtt{F}}
\newcommand{\G}{\mathtt{G}}
\newcommand{\Q}{\mathtt{Q}}
\newcommand{\K}{\mathtt{K}}
\renewcommand{\H}{\mathtt{H}}
\newcommand{\U}{\mathtt{U}}

\renewcommand{\t}{\mathbf{t}}
\newcommand{\R}{\mathtt{R}}

\begin{document}

\title{EKFPnP: Extended Kalman Filter for Camera Pose Estimation in a Sequence of Images\thanks{This paper is under review}}%
\date{2020\\ April}
\author{Mohammad Amin Mehralian \\
	Iran University of Science \& Technology\\
	\texttt{ma.mehralian@gmail.com}
	\and 
	Mohsen Soryani \\
	Iran University of Science \& Technology\\
	\texttt{soryani@iust.ac.ir}
}

\maketitle

\begin{abstract}
%%%
In real-world applications the Perspective-n-Point (PnP) problem should generally be applied in a sequence of images which a set of drift-prone features are tracked over time.
In this paper, we consider both the temporal dependency of camera poses and the uncertainty of features for the sequential camera pose estimation.
Using the Extended Kalman Filter (EKF), a priori estimate of the camera pose is calculated from the camera motion model and then corrected by minimizing the reprojection error of the reference points.
Experimental results, using both synthetic and real data, demonstrate that the proposed method improves the robustness of the camera pose estimation, in the presence of tracking error, compared to the state-of-the-art.
%%%%
\\[5pt]
\end{abstract}

%% main text
\section{Introduction}

Camera pose estimation by tracking a 3D object in a video sequence which is known as \textbf{3D tracking} means continuously identifying camera position and orientation relative to the scene, or, equivalently, the 3D displacement of an object relative to the camera when either the object or the camera is moving \cite{CGV-001}.
It has many applications in computer vision and robotics.
Examples include augmented reality, visual servoing of robots, incremental structure-from-motion, robot navigation, etc.
Generally, researchers have concentrated on estimating camera projection matrix from a set of 3D points in a reference world coordinate and their projection into the camera image coordinate.

Perhaps the simplest algorithm of this family is Direct Linear Transformation (DLT) which estimates the camera projection matrix by minimizing the 3D geometric error in the object space \cite{sutherland_1963,faugeras_1993,hartley_multiple_2004}.
Basic DLT calculates the most general form of the camera projection matrix with 11 degrees of freedom.
Thus, it is more fit for uncalibrated camera scenarios, in which case the problem is known as camera resectioning.
Although this algorithm can be adapted to more restricted camera models like affine model \cite{hartley_multiple_2004}, Ignoring the intrinsic parameters leads to inaccuracies when the camera is calibrated.

The second class of methods is Perspective-n-Point (PnP) techniques in which the calibration parameters are known.
In its simplest form, a closed-form solution is obtained from 3 noncollinear point correspondences (P3P) \cite{Fischler:1981:RSC:358669.358692,139759,67632} producing up to four possible solutions.
Generally, one additional correspondence is used to select the correct solution.

In real-world applications the PnP problem is usually applied to an image sequence.
However, to the authors' knowledge, most PnP solutions compute the camera pose independently from its history.
It means that they ignore the time dependency of camera poses through the image sequence.
This may cause instability in the estimation of camera motion sequence, especially when the observation noise level increases over time.

In this paper, the PnP problem is considered as a probabilistic estimation process over time.
It is assumed that the 3D reference points are fixed and the camera moves smoothly with small accelerations.
For this purpose we used Extended Kalman Filter (EKF).
EKF is one of the most popular recursive probabilistic filters which estimates the state of nonlinear systems.
The new state of the camera is predicted from its motion history and then corrected by minimizing the reprojection error of the reference points.
Therefore, the camera state contains the camera pose parameters, linear and angular velocities which are used to predict the camera pose over time.

Using the probabilistic approach, the uncertainties of observations (which may be caused by uncertainties of 3D reference points or 2D tracked points) will be considered in the problem formulation.
Other probabilistic methods such as CEPPnP \cite{ceppnp_2014} take into account a limiting assumption that the observation uncertainties are known in advance. We ignore this assumption since prior information of observation uncertainties may not be available.
By employing the EKF, final results are calculated by combining both camera motion model and reprojection error minimization.
Therefore, only a coarse estimate of observation uncertainties is enough to yield accurate results.

The method can evaluate the uncertainty of the pose parameters by estimating the camera state covariance.
This would be helpful when a quantitative criterion for measuring the accuracy of each parameter is needed.
Many solutions to this problem use image-space error, object-space error or other algebraic error functions as measures of pose reliability.
However, they cannot determine the amount of reliability for each camera pose parameter individually.

Moreover, by using EKF, additional sources of information such as inertial sensors can be combined to refine the estimation.
Additionally, the prediction step can be used to enhance other components of the algorithm such as feature tracking.

The paper proceeds as follows: Section 2 reviews the related work.
Section 3 describes our EKF formulation.
In section 4, experimental results are presented comparing our method with the state-of-the-art.
Section 5 concludes the paper.

\section{Related works}
P3P \cite{Fischler:1981:RSC:358669.358692} is one of the first algorithms for the pose estimation of calibrated cameras from 3 correspondences between 3D reference points and their 2D projections. 
In addition, there is a variety of PnP algorithms deriving closed-form solutions from a limited number of points, namely P4P \cite{Fischler:1981:RSC:358669.358692} and P5P \cite{triggs1999}.
Since these algorithms are only applicable to a small portion of correspondences, they are sensitive to noise.
Although other traditional PnP solutions have no restriction on the number of points, they are computationally expensive.
For example, the time complexity of \cite{fiore_2001} as the lowest complexity method is $O(n^2)$ while it is very sensitive to noise.
The time complexity of more robust methods like \cite{quan_1999} and \cite{ansar_2003} are significantly increased to $O(n^5)$ and $O(n^8)$ respectively.

Lepetit et al. introduced Efficient PnP (EPnP) \cite{epnp_2008} which is the first efficient non-iterative O(n) solution.
EPnP represents reference points by a weighted sum of four virtual control points.
Then the problem is solved using fourth order polynomials with simple linearization techniques.
The Robust PnP (RPnP) \cite{rpnp_2012} divides reference points into 3-point subsets in order to generate quadratic polynomials for each subset, and then the squared sum of those polynomials is used as a cost function.

In the Direct-Least-Squares (DLS) \cite{dls_2011} method, the PnP problem is solved by minimizing a nonlinear geometric cost function.
However, it suffers from rotational degeneracy since the Cayley representation is used for the rotations.
The Accurate and Scalable PnP (ASPnP) \cite{aspnp_2013} and the Optimal PnP (OPnP)\cite{opnp_2013} use a quaternion representation of rotation to overcome this problem and yield more accurate results.

All the mentioned methods are categorized as non-iterative methods in PnP.
The iterative approaches formulate the problem as a nonlinear least squares problem.
Although iterative algorithms are more robust to outliers and are more accurate, their performance highly depends on good initialization, since the objective function may be trapped in a local minimum.
Furthermore, they are computationally more expensive.
Some non-iterative methods use iterative algorithms like Gauss-Newton (GN) as a fine-tuner to achieve more accurate results.
For example, EPnP used this approach and is commonly known as EPnP+GN in the literature.

One of the fastest iterative PnP algorithms is the LHM method \cite{lhm_2000}.
It minimizes an error metric based on collinearity in object space and relies on an initial estimation of the camera pose with a weak-perspective assumption.
In contrast to LHM, the Procrustes PnP \cite{ppnp_2012} iteratively minimizes the error between the object and the back-projected image points.

With the possibility of outliers, it is necessary to incorporate an outlier-removal scheme like RANSAC \cite{Fischler:1981:RSC:358669.358692}.
Ferraz et al.
combine an algebraic outlier rejection strategy with the linear formulation of the PnP solution in EPnP algorithm called Robust Efficient Procrustes PnP (REPPnP)\cite{reppnp_2014}.
It sequentially removes correspondences yielding algebraic errors more than a specific threshold.
Final results are obtained by iteratively solving the closed-form Orthogonal Procrustes problem.
The authors have extended their method to integrate image points uncertainties introducing the Covariant EPPnP (CEPPnP) \cite{ceppnp_2014}.
To incorporate feature uncertainties in EPnP, a Gaussian distribution models the error for each of the observed 2D feature points.
Then the PnP solution is formulated as a maximum likelihood problem, approximated by an unconstrained Sampson error function.
It naturally penalizes the noisiest correspondences.
However, as noted in the article, the feature uncertainties are assumed to be known in advance.

It is also worth mentioning that some algorithms like EPnP, REPPnP and CEPPnP propose separate solutions for planar and non-planar reference points.
As a result, these methods may yield inaccurate results in cases with near-planar configurations \cite{opnp_2013}.

MLPNP \cite{mlpnp_2016} uses image points uncertainties to present a new maximum likelihood solution to the PnP problem.
First, the uncertainties propagate to the forward-projected bearing vectors.
Then the null space of bearing vectors is used to obtain a linear maximum likelihood solution.
Finally, the result of the ML estimator is iteratively refined with the Gauss-Newton optimization.

There are some related studies in the field of visual servoing which use the EKF for camera pose estimation;  for example, \cite{DONG2015291} and \cite{5560877} use EKF and Iterative Adaptive EKF (IAEKF) respectively for real-time control of robot motion.
Similar to our method, they formulate the control error in the image space.
However, they use the Euler angle representation of the rotation matrix.
Apart from the infamous gimbal lock problem, this adds to computational complexity of the algorithm \cite{shuster93}.

This paper develops a probabilistic approach to estimate camera poses in a sequence of images without use of any other source of information such as IMU.
For this purpose, the uncertainty of both 2D image points and 3D reference points are considered.
Additionally, the history of camera motion is also used to calculate an initial estimation of the camera pose.
To combine the uncertainty in the observed image points and the camera pose dynamics, Extended Kalman Filter (EKF) is applied recursively over time.

\section{Problem formulation}
Throughout this paper, matrices, vectors and scalars are denoted by capital letters, bold lowercase letters and plain lowercase letters, respectively. 

\subsection{Extended Kalman Filter}
Kalman filter is a powerful tool for recursively estimating the state of a linear dynamic process from a series of noisy measurements.
For non-linear systems Extended Kalman Filter (EKF) is widely used.
The EKF approximates a nonlinear model by its first order Taylor expansion.
Each step of EKF runs in two phases: \textit{prediction} and \textit{correction}.

The \textit{prediction} phase is independent of the current observation and only employs the dynamic model of the process to calculate a prior estimation of the current state from the posterior estimation of the previous state using the following equation:
\begin{equation}\label{eq:ekf_pre_s}
	\s_{k|k-1} = f(\s_{k-1|k-1}, \u_k),
\end{equation}
where $\s$ is the state vector and $f(.)$ is the nonlinear dynamic model of the process.
The term $\u_k$ is the dynamic model noise vector which is independent of the state vector and is commonly distributed with zero mean.

The uncertainty of this prediction can be represented by the covariance matrix $\P_{k|k-1}$:
\begin{equation}\label{eq:ekf_pre_p}
	\P_{k|k-1}=\F_{k-1} \P_{k-1|k-1} \F_{k-1}^T+\G_{k-1} \Q_{k-1} \G_{k-1}^T,
\end{equation}
where $\F_k$ is the derivative of the dynamic model with respect to the state vector $\s$, $\G_k$ is the derivative of the dynamic model with respect to $\u_k$, and $\Q_k$ is the noise covariance.

In the \textit{correction} phase, the prior estimation of the current state is corrected to the posterior estimation given a new observation.
By incorporating the observations, the posterior of the state and its covariance matrix can be estimated as
\begin{align}
	\s_{k|k} &=\s_{k|k-1}+\K_{k}(\z_k-h(\s_{k|k-1})),   \label{eq:ekf_cor_s}
	\\
	\P_{k|k} &=(I-\K_{k} \H_{k}) \P_{k|k-1},	\label{eq:ekf_cor_p}
\end{align}
where $\z_k$ represents the observations, $h(.)$ is the observation model mapping the state space into the observed space $\z_k$ and $\H_k$ is the derivative of $h(.)$ with respect to the state vector $\s$.
The parameter $\K$, known as the Kalman gain, specifies how much the posterior should be affected by observation error and is computed as
\begin{equation}\label{eq:ekf_cor_k}
	\K_{k}= \P_{k|k-1}\H_k^T (\H_k \P_{k|k-1} \H_k^T+\U_k)^{-1},
\end{equation}
where $\U$ is the observation noise covariance.

To apply EKF to sequential camera pose estimation, our formulation closely follows the approach of MonoSLAM \cite{mono_slam_2007}.
However, contrary to our approach, MonoSLAM puts the 3D reference points in its state vector which greatly increases the amount of computation due to the very large size of matrices and vectors.
%MonoSLAM order becomes to $O(N^2)$  where N is the number of features
%This means that the number of features which can be maintained with realtime processing is bounded in MonoSLAM.

\subsection{State vector and motion model}
In general, the projection of a perspective camera is defined by its translation vector $\t$ and rotation matrix $\R$ with respect to the world reference coordinates.
This transformation for a 3D point $\x^w=[x_i,x_j,x_k ]^T$ is shown by
\begin{equation}\label{eq:projection_t}
	\x^c=\R\x^w+\t.
\end{equation}
This equation can be expressed with respect to the camera center $\c$:
\begin{equation}\label{eq:projection_c}
	\x^c=\R(\x^w-\c).
\end{equation}

It is known that the Rotation matrix should obey both the orthonormal constraint $\R\R^T=I$ and the determinant constraint $det(\R) = 1$.
One of the most convenient ways to deal with these constraints is to use the quaternion representation.
Furthermore, as a four-element vector, it is more compact than the rotation matrix representation.

Accordingly, the state vector of EKF includes camera center position $\c$, camera orientation $\q$, and both linear and angular velocities $\v$ and $\w$:
\begin{equation}
	\s_{13\times1} = [\c_{3\times1}^T, \q_{4\times1}^T, \v_{3\times1}^T, \w_{3\times1}^T]^T.
\end{equation}

For the dynamic model, the camera motion is considered smooth with a constant velocity.
In fact both linear and angular accelerations are modeled with zero-mean Gaussian noise  $\w=[a, \alpha]^T$. Thus the dynamic model can be written as:
\begin{equation}
	f_{13\times1}(\s,n)=
	\begin{bmatrix}
		\c_k+ (\v_k+\mathbf{\nu}_k) \dt\\
		\q_k * \q( (\w_k + \mathbf{\Omega}_k) \dt) \\
		\v_k + \mathbf{\nu}_k\\
		\w_k + \mathbf{\Omega}_k
	\end{bmatrix},
\end{equation}
where $\mathbf{\nu}=a \dt$ and $\mathbf{\Omega}=\alpha \dt$ are the velocities caused by linear and angular accelerations, and $q((\w_k+\mathbf{\Omega})\dt)$ is the unit quaternion equivalent of angle $(\w_k+\mathbf{\Omega})\dt$.
The symbol "*" means quaternion product.

To calculate the state prediction covariance of Eq. \eqref{eq:ekf_pre_p} it is necessary to calculate the Jacobian of $f(.)$ with respect to $\s$ and $\w$:
\begin{align}
	\F_{13\times13} &=\begin{bmatrix}
		I_3 & 0 & \dt I_3 & 0 \\
		0   & \frac{\delta \q_{k+1}}{\delta \q_k} & 0 & \frac{\delta \q_{k+1}}{\delta \w_k} \\
		0 & 0 & I_3 & 0\\
		0 & 0 & 0 & I_3
	\end{bmatrix},
	\\
	\G_{13\times6} &=
	\begin{bmatrix}
		\dt I_3 & 0 \\
		0   & \frac{\delta \q_{k+1}}{\delta \mathbf{\Omega}_k} \\
		I_3 & 0\\
		0 & I_3
	\end{bmatrix}.
\end{align}
To see how the derivatives in the quaternion space are calculated please refer to \cite{civera2012structure}

\subsection{Observation model}
\begin{figure}[!t]
	\centering
	\includegraphics[scale=0.9]{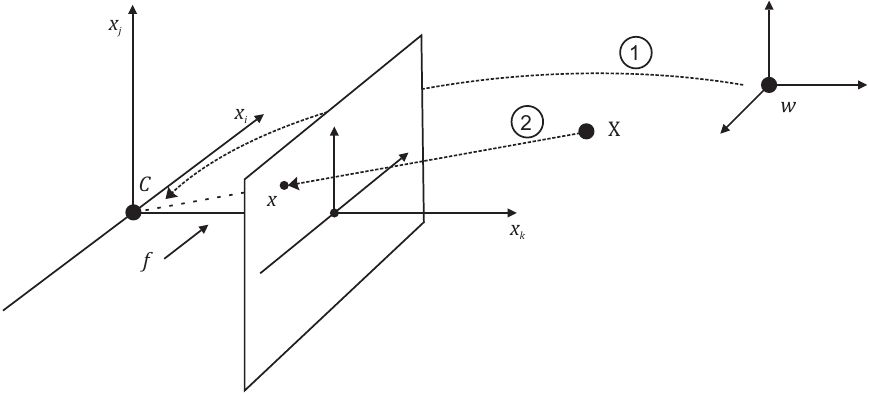}
	\caption{Two steps of the observation model which maps 3D world reference points to 2D image points}
	\label{fig:projection}
\end{figure}

For EKF-based PnP, observations are 2D correspondences of 3D reference points in the world coordinate, acquired by the moving camera.
So, for $n$ 2D points the observations vector is
\begin{equation}
	\z_{2n \times 1} = [x^1_i, x^1_j, \cdots, x^n_i, x^n_j]^T.
\end{equation}
The observation model is the projection function mapping the 3D points into the image plane.
This mapping is typically done in two steps (Fig. \ref{fig:projection}).

First, 3D reference points are transformed from the world coordinate system to the camera coordinate system using Eq. \eqref{eq:projection_c}. Eq. \eqref{eq:projection_c} is rewritten in terms of quaternions as 
\begin{equation}\label{eq:projection_q}
	h^{WC}(\s, \mathbf{x}^w) =
	\begin{bmatrix}
		0 \\ \x^c
	\end{bmatrix} =
	\q*
	\begin{bmatrix}
		0 \\ \x^w-\c
	\end{bmatrix}*
	\mathbf{q^*},
\end{equation}
where $\q$ is the unit-quaternion equivalent to the rotation matrix and $\q^*$ is the conjugate of $\q$.

Second, 3D points are transformed from the camera coordinate system to the image plane using pinhole camera model:
\begin{equation}\label{eq:projection_f}
	h^{CI}(\x^c) = \x^I = 
	[f\frac{x_i^c}{x_k^c},f\frac{x_j^c}{x_k^c } ].
\end{equation}
where $f$ is the camera focal length.
Thus the observation model is
\begin{equation}
	h_{2n \times 1}=h^{CI} (h^{WC}(\x^w)).
\end{equation}

\subsubsection{Kalman gain}
Compared to the usual PnP methods, our method is able to reduce the effect of the observation noise, as the dynamic model based prediction and the observation-based correction steps are combined using the Kalman gain to offer a more robust estimation. 

The observation covariance matrix $\U$ of Eq. \eqref{eq:ekf_cor_k} is the uncertainty of the tracked 2D points in the image sequence.
It is assumed that feature tracking errors are independent and identically distributed with a zero-mean Gaussian distribution.
Thus, $\U$ is diagonal.
In some literatures like CEPPnP \cite{ceppnp_2014} and MLPnP \cite{mlpnp_2016}, it is common to assume that the observation covariance is known a priori, although this is hardly the case in real applications.
Our method can handle this issue as the  EKF  is able to work with an overestimation of the covariance. 

To estimate the Kalman gain and the posterior state covariance matrix, the Jacobian of observation model $h$ with respect to the state vector $\s$ can be derived according to 
\begin{equation}
	\H_{2n \times 13} = \frac{\delta h}{\delta \s} = \frac{\delta h^{CI}}{\delta h^{WC}} \frac{\delta h^{WC}}{\delta \s}.
\end{equation}
For details please refer to \cite{civera2012structure}

\section{Experimental results}
We compare the proposed algorithm (EKFPnP) to the state-of-the-art, including EPnP+GN \cite{epnp_2008}, DLS \cite{dls_2011}, PPnP \cite{ppnp_2012}, RPnP \cite{rpnp_2012}, ASPnP \cite{aspnp_2013}, OPnP \cite{opnp_2013}, CEPPnP \cite{ceppnp_2014} and MLPnP \cite{mlpnp_2016},
using both synthetic and real data.
We use the MATLAB implementation of the other methods, and made our own data and implementation available at \url{https://github.com/ma-mehralian/ekfpnp_matlab_toolbox}.

\subsection{Simulated Data }
\begin{figure}[t]
	\centering
	\includegraphics[scale=0.7, trim={1cm 0.9cm 0 1cm},clip]{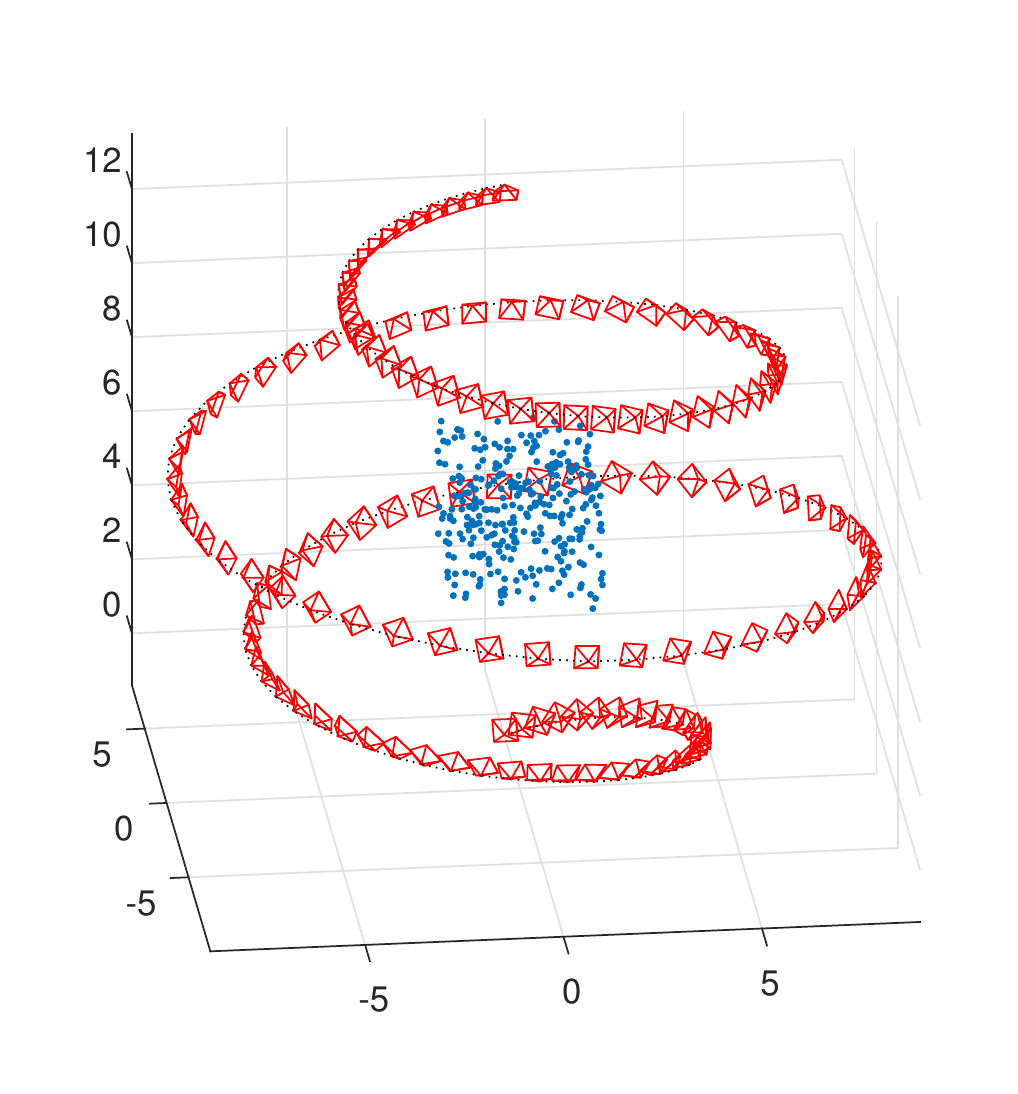}
	\caption{The camera trajectory for generation of simulated data}
	\label{fig:cam_trajectory}
\end{figure}

In this section, the accuracy and execution time of EKFPnP are compared with the other methods.
Our set up is very similar to that of CEPPnP and MLPnP.
All simulations were repeated 100 times independently, and the average errors are reported.
The rotation error (degree) is computed as
\begin{equation}
	e_{rot} (deg)=max_{k=1}^3\{arccos(r_{k,true}^T.r_k )\times 180/\pi\},
\end{equation}
where $r_{k,true}^T$ and $r_k$ are the k-th columns of $R_{true}$ and $R$ respectively.
Also, the translation error (\%) is defined as
\begin{equation}
	e_{trans} (\%)=  \|t_{true}-t\|/\|t\| \times 100.
\end{equation}
We generate both non-planar and planar 3D data.
In the first setting, the non-planar 3D points were uniformly sampled from the interval $x, y, z \in [-2,2]\times[-2,2]\times[4,8]$.
In the second setting the planar 3D points were generated by uniformly sampling $x, y\in [-2,2]\times[-2,2]$ and setting $z=0$.
As shown in Fig. \ref{fig:cam_trajectory} the image of the 3D points are taken by a camera moving along a trajectory with 200 virtual camera poses.
The camera is moved in such a way that all extrinsic parameters are changed in each step of the trajectory.
For all virtual perspective cameras, the focal length is set to 800 pixels, and we assumed an image size of $640 \times 480$.

\begin{figure*}[b]
	\centering
	\includegraphics[width=0.8\textwidth]{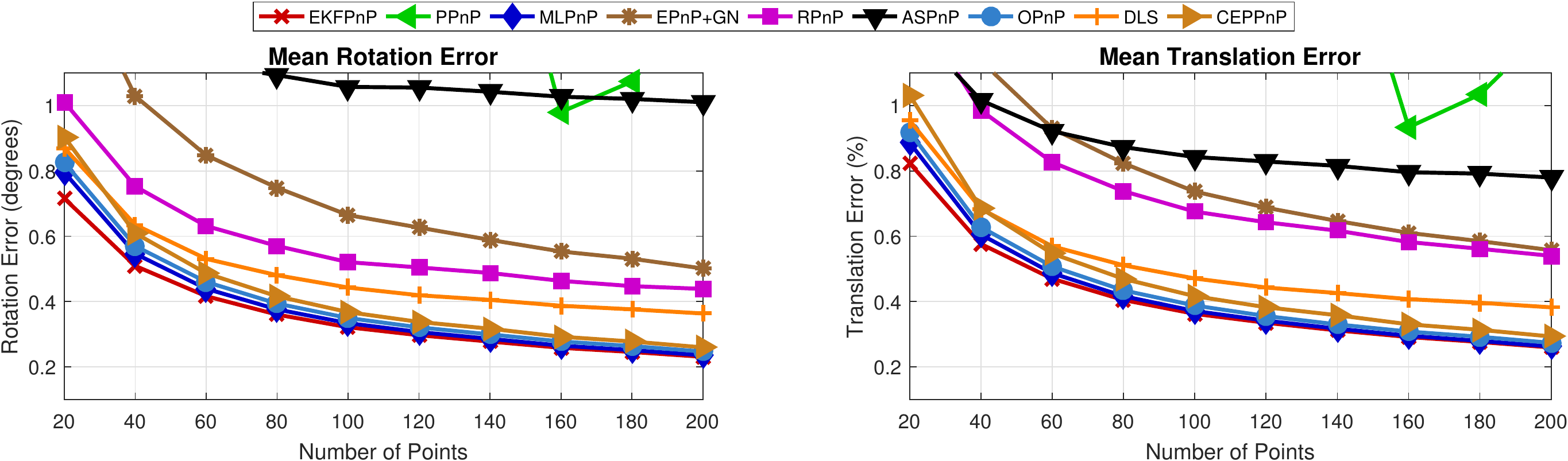}
	\\ \vspace{0.5em}
	\includegraphics[width=0.8\textwidth]{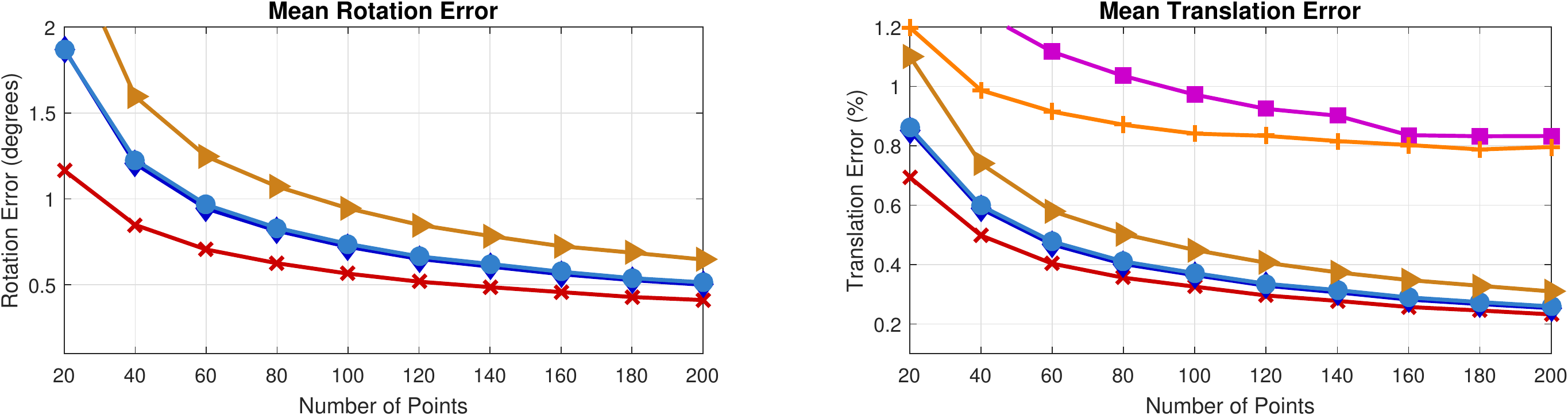}
	\vspace{-0.5em}
	\caption{Mean translation and rotation errors for varying point numbers. First row: ordinary-3D points, second row: planar-3D points.}
	\label{fig:expr_sim_vp}
\end{figure*}

In the first experiment, the competing algorithms are evaluated with different noise levels.
Every 10\% of 2D image points are perturbed by one of 10 zero-mean Gaussian noise distributions with different standard deviations of $\sigma = [1...10]$. To make the observation noise more realistic, the magnitude of noise is increased gradually over time from 0\% to 100\%.
CEPPnP, MLPnP and the proposed method use 2D points uncertainty as part of the algorithm's input.
We do not use the variance of the observations known for the simulated data.
Instead, we provide the algorithms with an overestimation of the variance.
Thus, for these 3 algorithms the average value of points' variances is used as a common variance for all points.
For the proposed method, the velocity of the camera in the state vector of the Kalman filter should be initialized.
This is done by estimating the camera pose in the first pair of images using EPnP+GN.

Fig. \ref{fig:expr_sim_vp} compares the accuracy of the proposed method with other methods for different numbers of correspondences.
The first and second rows show the average rotation and translation errors for n=20 to 200 correspondences for the non-planar and planar settings, respectively. 

It can be seen that the proposed algorithm outperforms the other methods in terms of the estimation error, especially in the planar case.
As expected, since the observation uncertainties are taken into account in the EKFPnP, MLPnP and CEPnP methods, they produce competitive results. 

\begin{figure*}[!t]
	\centering
	\includegraphics[width=0.9\textwidth]{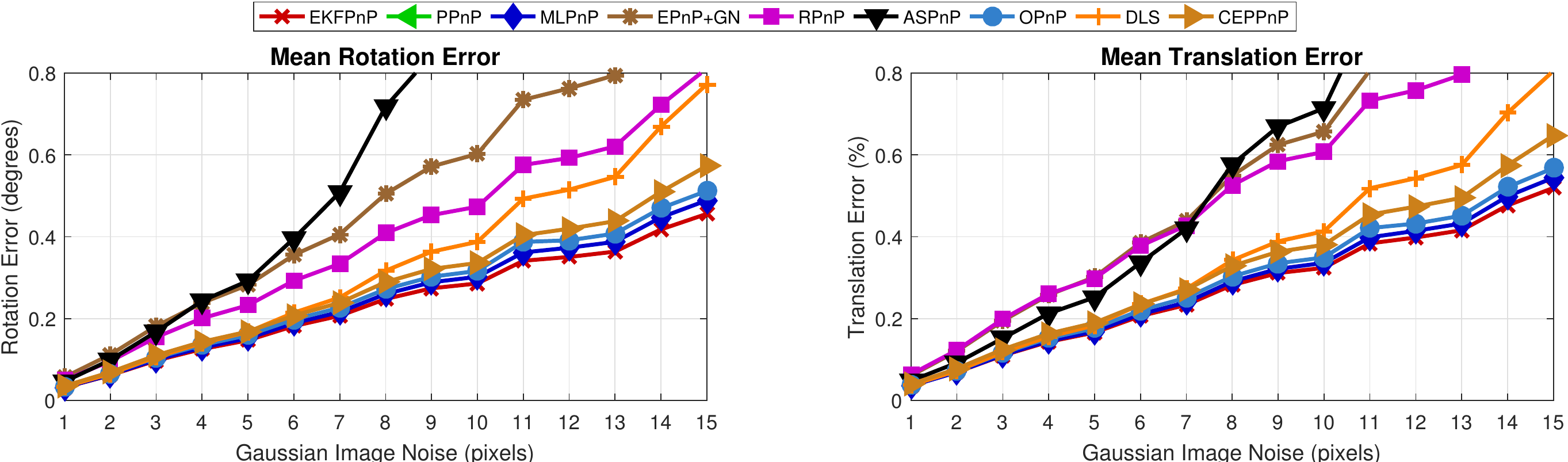}\\
	\vspace{1em}
	\includegraphics[width=0.9\textwidth]{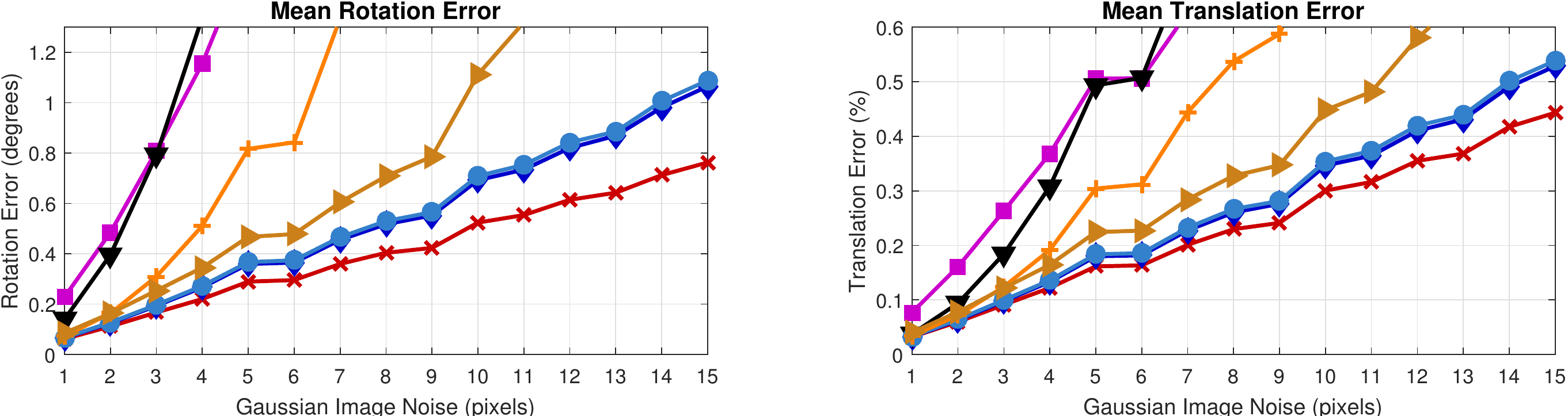}
	\caption{Mean translation and rotation errors for varying noise levels. First row: ordinary-3D points, second row: planar-3D points.}
	\label{fig:expr_sim_vn}
\end{figure*}

In the second experiment, it is assumed that the number of correspondences remains fixed with n=100 and the noise level varies.
Fig. \ref{fig:expr_sim_vn} shows the results obtained by ranging the noise variance from 1 to 15.
Again the magnitude of noise is increased gradually over time.
It can be seen that the proposed method still maintains its superiority over other methods as the noise level increases. 

\begin{figure}[!h]
	\centering
	\includegraphics[width=0.9\linewidth]{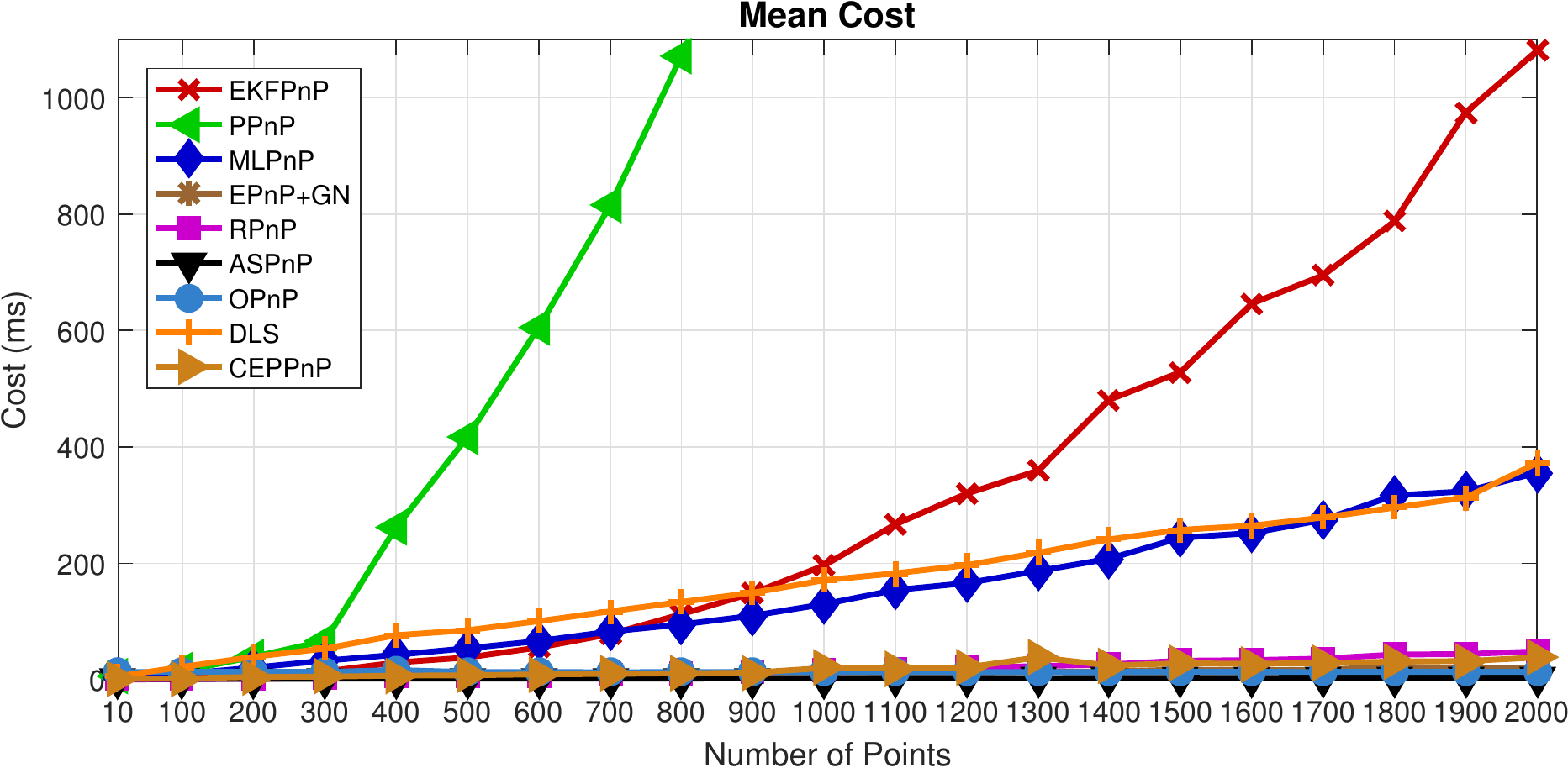}
	\caption{Comparison of runtime of all methods for different number of points.}
	\label{fig:expr_sim_t}
\end{figure}

It is hard to compare our runtime performance with other algorithms since our MATLAB implementation  is not optimized.
Moreover, some other algorithms have improved their performance by using a C++ implementation plus MATLAB MEX.
Fig. \ref{fig:expr_sim_t} depicts the running times of all algorithms as the number of points increases.  

\subsection{Real Data}
In these experiments, the accuracy of the proposed method is evaluated on real image sequences.
In the first experiment camera pose or, equivalently, the 3D displacement of a box relative to the camera is tracked in the sequence.
The second experiment employs camera pose estimation in an incremental structure-from-motion system.
CEPPnP \cite{ceppnp_2014}, MLPnP \cite{mlpnp_2016} and EKFPnP, which are the algorithms that consider uncertainties in the observed feature points where applied to above data and results were compared. 

\subsubsection{Box tracking}

In this experiment data consists of a sequence of 573 images of a freely moving box with known dimensions.
All frames have a resolution of $1280 \times 720$ pixels and the camera was calibrated using Camera Calibration Toolbox for Matlab \cite{bouguet_2008}.
The 3D model of the box including 898 ORB features \cite{rublee_2011} was created.
To track the features, Kanade-Lucas-Tomasi (KLT) \cite{tomasi91} tracker was employed.
As result using 3D model and 2D tracked features we estimate the camera pose and the 3D motion of the box.

The 2D features are tracked in images and used as observations.
In the tracking phase, the tracked points are usually corrupted over time by tracking errors or they are even get lost due to occlusions or image brightness inconsistency.
Thus, this experiment is a good way to evaluate algorithms' robustness against tracking errors.

\begin{figure*}[!h]
	\centering
	\includegraphics[width=0.8\textwidth]{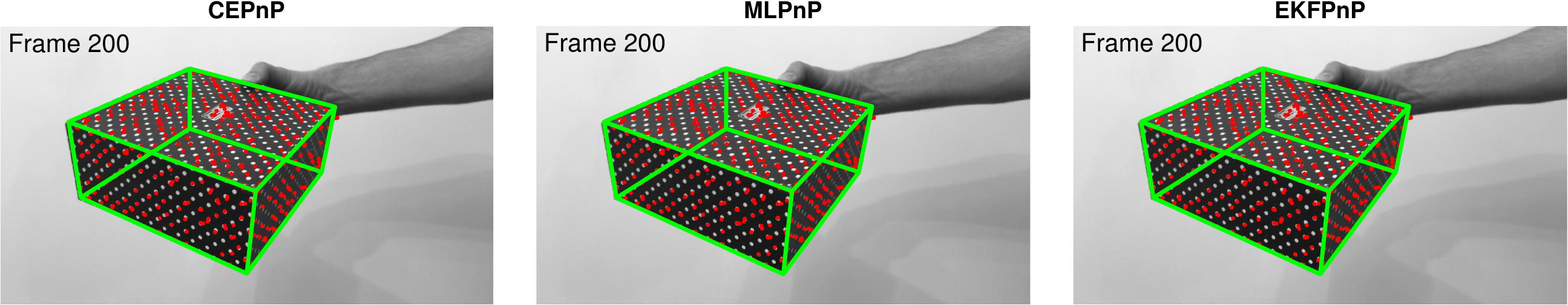}
	\includegraphics[width=0.8\textwidth]{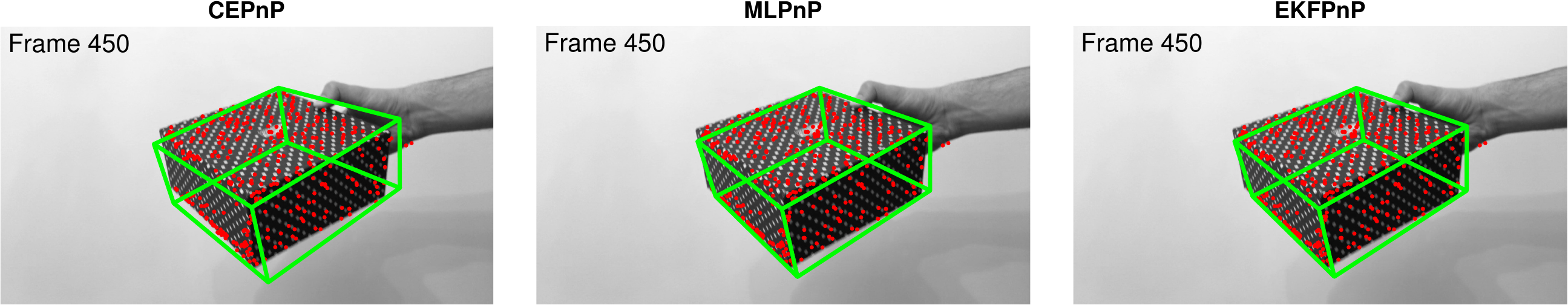}
	\includegraphics[width=0.8\textwidth]{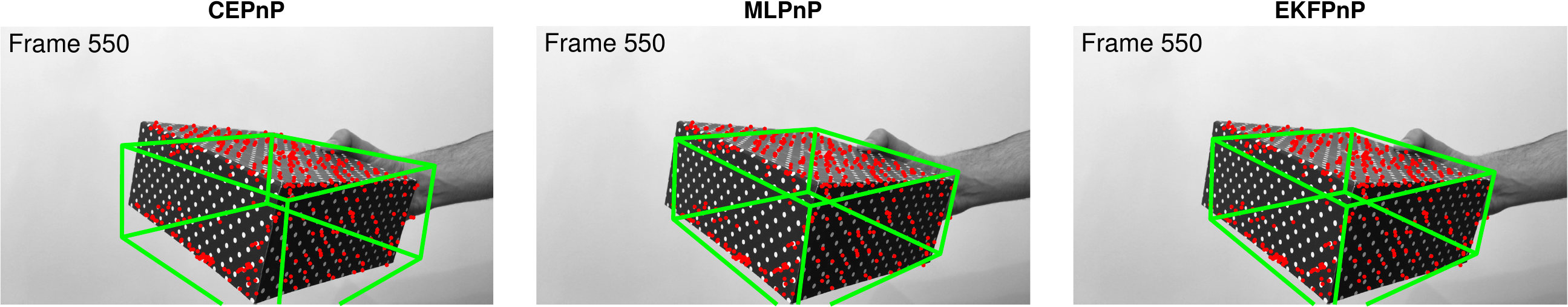}	
	\caption{Experimental results of the box pose estimation using CEPPnP, MLPnP and EKFPnP. First, second and third row shows algorithms' results for frames 200, 450 and 550 of the composed video respectively.}
	\label{fig:expr_real_box}
\end{figure*}

The results obtained using each method are given in Fig. \ref{fig:expr_real_box}.
Since the exact amount of tracking error is not known, a constant approximate mean value of 3 pixels per points was used for CEPnP.
Clearly, our method outperforms CEPnP, while the results of MLPnP and EKFPnP are quite similar. 

\subsubsection{Structure from Motion}
\begin{figure}[!h]
	\centering
	\includegraphics[width=0.8\linewidth]{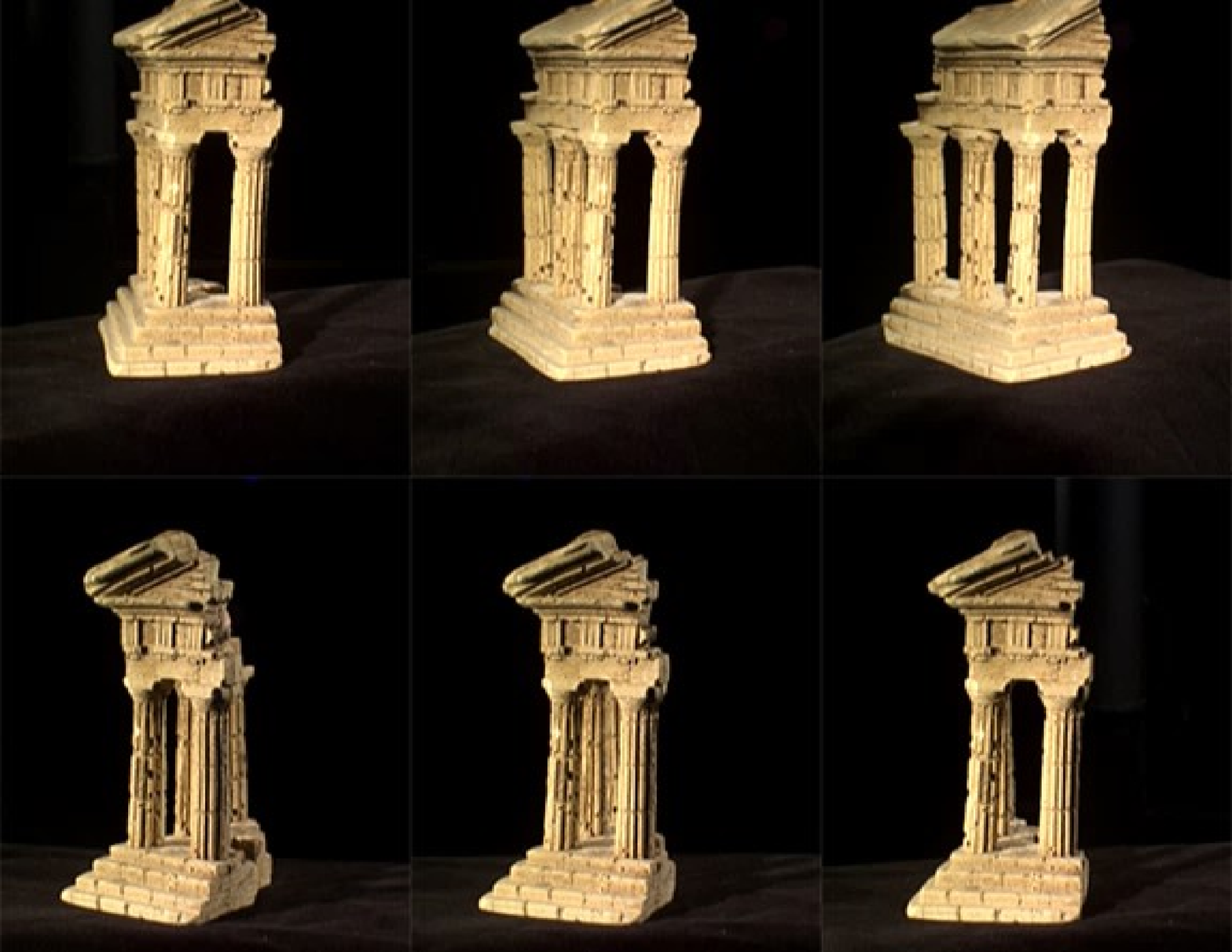}
	\caption{Sample images from TempleRing dataset}
	\label{fig:expr_real_temple}
\end{figure}

In this experiment we use part of TempleRing dataset \cite{temple_dataset} (Fig. \ref{fig:expr_real_temple}) to develop structure from motion.
The data consists of a sequence of 19, $640 \times 480$ images sampled on a ring around a plaster temple.

\begin{figure}[!h]
	\centering
	\includegraphics[width=\linewidth]{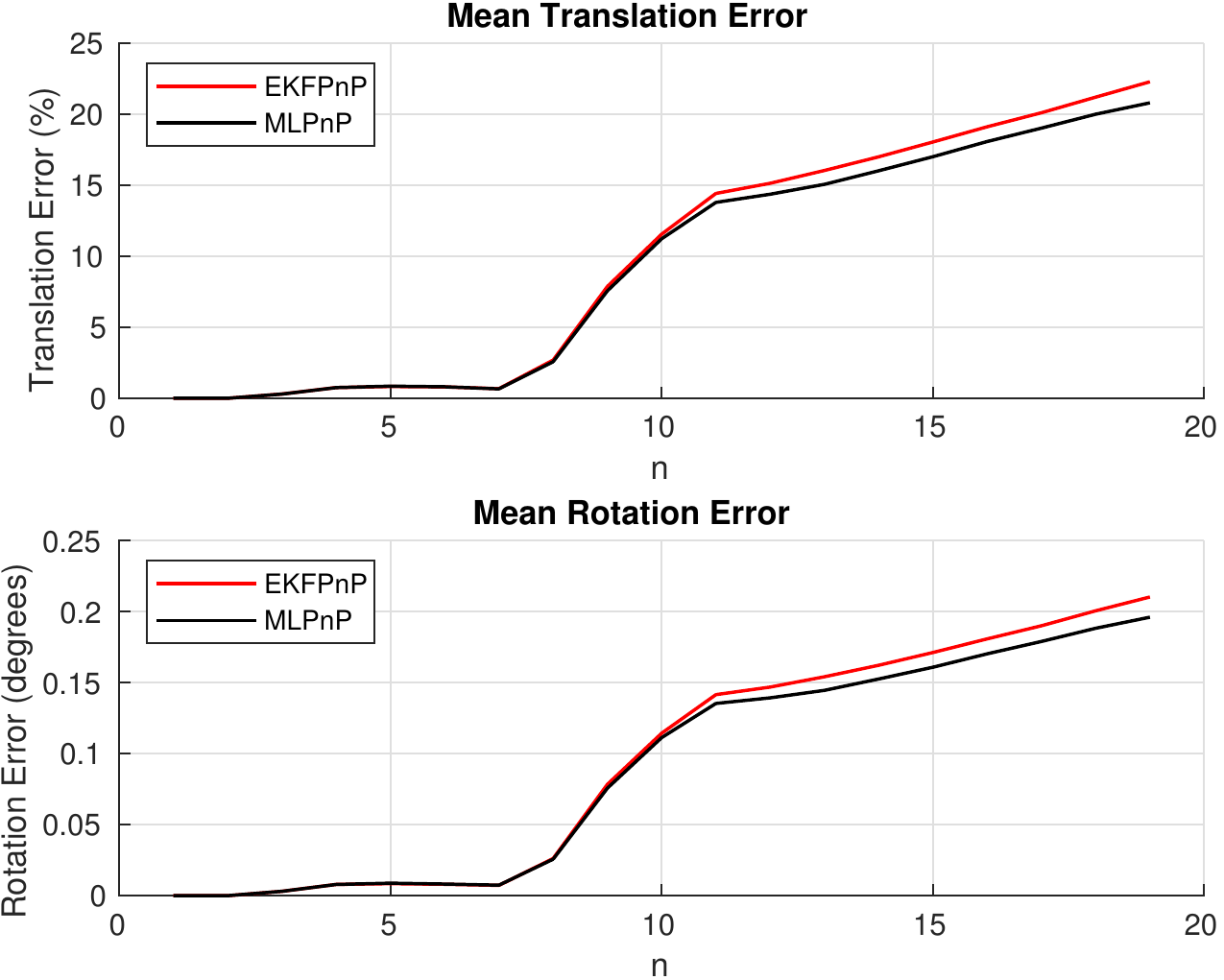}
	\caption{Mean translation and rotation errors for SFM experiment comparing MLPnP and EKFPnP methods. }
	\label{fig:expr_real_sfm}
\end{figure}

To estimate camera poses with the same scale of ground truth, we use first two poses from ground truth to initialize 3D model with triangulation of 2D features.
We extract SIFT features \cite{lowe2004} in each image and match them with the features of the previous view.
Using 2D features and 3D model correspondences we estimate camera poses for the rest of sequence. 
If 3D correspondences of some 2D features do not exist in 3D model we add them to the model by triangulation.
Since we use estimated poses to triangulate new 3D points, the error of both 3D model and pose estimation is increased gradually.
Fig. \ref{fig:expr_real_sfm} shows rotation and translation errors of MLPnP and EKFPnP.
Experiments were repeated 50 times independently, and the average errors are reported.
We have remove CEPnP from comparison because it produces large error values.
As can be seen, the results of both methods are very close. 

\begin{figure}[!h]
	\centering
	\includegraphics[width=0.9\linewidth,trim={4cm 4cm 3cm 3cm},clip]{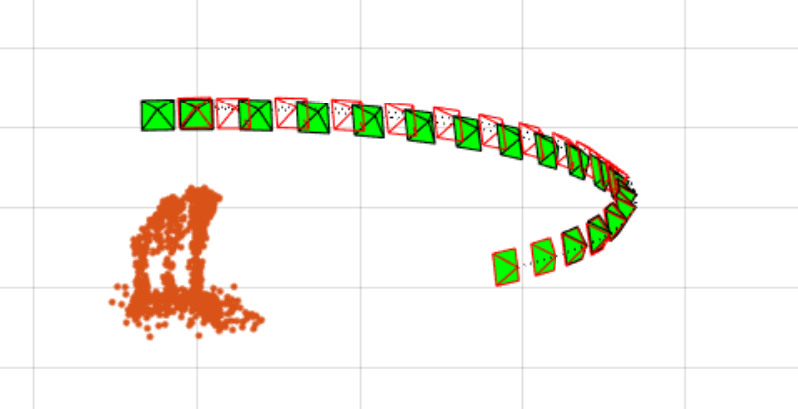}
	\caption{SFM results of EKFPnP method}
	\label{fig:expr_real_sfm_ekfpnp}
\end{figure}

\begin{figure}[!h]
	\centering
	\includegraphics[width=0.9\linewidth,trim={4cm 4cm 3cm 3cm},clip]{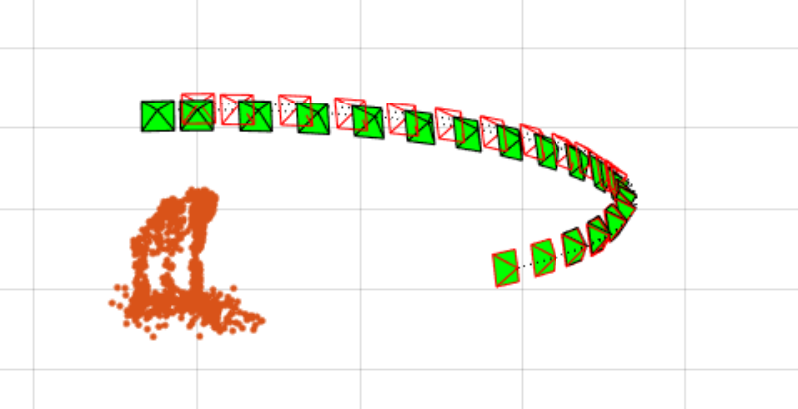}
	\caption{SFM results of MLPnP method}
	\label{fig:expr_real_sfm_mlpnp}
\end{figure}

Fig. \ref{fig:expr_real_sfm_ekfpnp} and Fig. \ref{fig:expr_real_sfm_mlpnp} show the results of SFM for EKFPnP and MLPnP respectively. In both figures the ground truth and estimated poses are shown in green red respectively.


\section{Conclusion}

In this paper we incorporated both camera motion history and 2D features' uncertainty to estimate camera pose in a sequence of images.
We employed an extended Kalman filter (EKF) as a recursive probabilistic model for estimation of the camera pose in two steps of prediction from the motion model and correction by minimization of the reprojection error.
This approach also provides the covariance of the pose estimated parameters which helps to measure the reliability of the results. 

Our synthetic and real data experiments revealed that the incorporation of 2D features' uncertainty and motion dynamics produces results which are more robust against feature tracking errors.

\bibliography{refs} 

\begin{thebibliography}{10}

\bibitem{CGV-001}
V.~Lepetit and P.~Fua, ``Monocular model-based 3d tracking of rigid objects: A
  survey,'' {\em Foundations and Trends® in Computer Graphics and Vision},
  vol.~1, no.~1, pp.~1--89, 2005.

\bibitem{sutherland_1963}
I.~Sutherland, ``Sketchpad: A man machine graphical communications system,''
  tech. rep., MIT Lincoln Laboratories, 1963.

\bibitem{faugeras_1993}
O.~Faugeras, {\em Three-Dimensional Computer Vision: A Geometric Viewpoint}.
\newblock MIT Press, 1993.

\bibitem{hartley_multiple_2004}
R.~I. Hartley and A.~Zisserman, {\em Multiple view geometry in computer
  vision}.
\newblock Cambridge, UK ; New York: Cambridge University Press, 2nd~ed., 2004.

\bibitem{Fischler:1981:RSC:358669.358692}
M.~A. Fischler and R.~C. Bolles, ``Random sample consensus: A paradigm for
  model fitting with applications to image analysis and automated
  cartography,'' {\em Commun. ACM}, vol.~24, pp.~381--395, June 1981.

\bibitem{139759}
R.~M. Haralick, D.~Lee, K.~Ottenburg, and M.~Nolle, ``Analysis and solutions of
  the three point perspective pose estimation problem,'' in {\em Proceedings.
  1991 IEEE Computer Society Conference on Computer Vision and Pattern
  Recognition}, pp.~592--598, Jun 1991.

\bibitem{67632}
W.~J. Wolfe, D.~Mathis, C.~W. Sklair, and M.~Magee, ``The perspective view of
  three points,'' {\em IEEE Transactions on Pattern Analysis and Machine
  Intelligence}, vol.~13, pp.~66--73, Jan 1991.

\bibitem{ceppnp_2014}
L.~Ferraz, X.~Binefa, and F.~Moreno-Noguer, ``Leveraging feature uncertainty in
  the pnp problem,'' in {\em Proceedings of the British Machine Vision
  Conference}, BMVA Press, 2014.

\bibitem{triggs1999}
B.~Triggs, ``Camera pose and calibration from 4 or 5 known 3d points,'' in {\em
  Proceedings of the Seventh IEEE International Conference on Computer Vision},
  vol.~1, pp.~278--284 vol.1, 1999.

\bibitem{fiore_2001}
P.~D. Fiore, ``Efficient linear solution of exterior orientation,'' {\em IEEE
  Transactions on Pattern Analysis and Machine Intelligence}, vol.~23,
  pp.~140--148, Feb 2001.

\bibitem{quan_1999}
L.~Quan and Z.~Lan, ``Linear n-point camera pose determination,'' {\em IEEE
  Transactions on Pattern Analysis and Machine Intelligence}, vol.~21,
  pp.~774--780, Aug 1999.

\bibitem{ansar_2003}
A.~Ansar and K.~Daniilidis, ``Linear pose estimation from points or lines,''
  {\em IEEE Transactions on Pattern Analysis and Machine Intelligence},
  vol.~25, pp.~578--589, May 2003.

\bibitem{epnp_2008}
V.~Lepetit, F.~Moreno-Noguer, and P.~Fua, ``Epnp: An accurate o(n) solution to
  the pnp problem,'' {\em International Journal of Computer Vision}, vol.~81,
  no.~2, pp.~155--166, 2008.

\bibitem{rpnp_2012}
S.~Li, C.~Xu, and M.~Xie, ``A robust o(n) solution to the perspective-n-point
  problem,'' {\em IEEE Transactions on Pattern Analysis and Machine
  Intelligence}, vol.~34, pp.~1444--1450, July 2012.

\bibitem{dls_2011}
J.~A. Hesch and S.~I. Roumeliotis, ``A direct least-squares (dls) method for
  pnp,'' in {\em 2011 International Conference on Computer Vision},
  pp.~383--390, Nov 2011.

\bibitem{aspnp_2013}
Y.~Zheng, S.~SUGIMOTO, and M.~OKUTOMI, ``Aspnp: An accurate and scalable
  solution to the perspective-n-point problem,'' {\em IEICE Transactions on
  Information and Systems}, vol.~E96.D, no.~7, pp.~1525--1535, 2013.

\bibitem{opnp_2013}
Y.~Zheng, Y.~Kuang, S.~Sugimoto, K.~Åström, and M.~Okutomi, ``Revisiting the
  pnp problem: A fast, general and optimal solution,'' in {\em 2013 IEEE
  International Conference on Computer Vision}, pp.~2344--2351, Dec 2013.

\bibitem{lhm_2000}
C.~P. Lu, G.~D. Hager, and E.~Mjolsness, ``Fast and globally convergent pose
  estimation from video images,'' {\em IEEE Transactions on Pattern Analysis
  and Machine Intelligence}, vol.~22, pp.~610--622, Jun 2000.

\bibitem{ppnp_2012}
V.~Garro, F.~Crosilla, and A.~Fusiello, ``Solving the pnp problem with
  anisotropic orthogonal procrustes analysis,'' in {\em 2012 Second
  International Conference on 3D Imaging, Modeling, Processing, Visualization
  Transmission}, pp.~262--269, Oct 2012.

\bibitem{reppnp_2014}
L.~Ferraz, X.~Binefa, and F.~Moreno-Noguer, ``Very fast solution to the pnp
  problem with algebraic outlier rejection,'' in {\em 2014 IEEE Conference on
  Computer Vision and Pattern Recognition}, pp.~501--508, June 2014.

\bibitem{mlpnp_2016}
S.~Urban, J.~Leitloff, and S.~Hinz, ``Mlpnp - a real-time maximum likelihood
  solution to the perspective-n-point problem,'' in {\em ISPRS Annals of
  Photogrammetry, Remote Sensing \& Spatial Information Sciences}, vol.~3,
  pp.~131--138, 2016.

\bibitem{DONG2015291}
G.~Dong and Z.~Zhu, ``Position-based visual servo control of autonomous robotic
  manipulators,'' {\em Acta Astronautica}, vol.~115, pp.~291 -- 302, 2015.

\bibitem{5560877}
F.~Janabi-Sharifi and M.~Marey, ``A kalman-filter-based method for pose
  estimation in visual servoing,'' {\em IEEE Transactions on Robotics},
  vol.~26, pp.~939--947, Oct 2010.

\bibitem{shuster93}
M.~D. Shuster, ``Survey of attitude representations,'' {\em Journal of the
  Astronautical Sciences}, vol.~41, pp.~439--517, Oct. 1993.

\bibitem{mono_slam_2007}
A.~Davison, I.~Reid, N.~Molton, and O.~Stasse, ``Monoslam: Real-time single
  camera slam,'' {\em IEEE Transactions on Pattern Analysis and Machine
  Intelligence}, vol.~29, pp.~1052--1067, June 2007.

\bibitem{civera2012structure}
J.~Civera, A.~J. Davison, and J.~M.~M. Montiel, {\em Structure from Motion
  Using the Extended Kalman Filter}.
\newblock Springer, 2012.

\bibitem{bouguet_2008}
J.-Y. Bouguet, ``Camera calibration toolbox for matlab,'' June 2008.

\bibitem{rublee_2011}
E.~Rublee, V.~Rabaud, K.~Konolige, and G.~Bradski, ``Orb: An efficient
  alternative to sift or surf,'' in {\em 2011 International Conference on
  Computer Vision}, pp.~2564--2571, Nov 2011.

\bibitem{tomasi91}
C.~Tomasi and T.~Kanade, ``Detection and tracking of point features,'' tech.
  rep., International Journal of Computer Vision, 1991.

\bibitem{temple_dataset}
D.~S. B.~C. Steve~Seitz, James~Diebel and R.~Szeliski, ``{TempleRing} data
  set.''

\bibitem{lowe2004}
D.~G. Lowe, ``Distinctive image features from scale-invariant keypoints,'' {\em
  International Journal of Computer Vision}, vol.~60, no.~2, pp.~91--110, 2004.

\end{thebibliography}
\bibliographystyle{ieeetr}

\end{document}